\documentclass[conference]{IEEEtran}
\IEEEoverridecommandlockouts
\usepackage{cite}
\usepackage{soul,color}
\usepackage{amsmath,amssymb,amsfonts}
\usepackage{algorithmic}
\usepackage{graphicx}
\usepackage{textcomp}
\usepackage{xcolor}
\def\BibTeX{{\rm B\kern-.05em{\sc i\kern-.025em b}\kern-.08em
    T\kern-.1667em\lower.7ex\hbox{E}\kern-.125emX}}
    
\usepackage[autostyle, english = american]{csquotes}
\MakeOuterQuote{"}    
\usepackage{fancyhdr}

\begin{document}

\title{Affective Robotics For Wellbeing: A Scoping Review
}

\author{\IEEEauthorblockN{Micol Spitale}
\IEEEauthorblockA{\textit{Department of Computer Science and Technology} \\
\textit{University of Cambridge}\\
Cambridge, UK \\
ms2871@cam.ac.uk}
\and
\IEEEauthorblockN{Hatice Gunes}
\IEEEauthorblockA{\textit{Department of Computer Science and Technology} \\
\textit{University of Cambridge}\\
Cambridge, UK \\
hg410@cam.ac.uk}
}

\maketitle

\thispagestyle{fancy}
\renewcommand{\headrulewidth}{0pt}
\fancyhf{}
\fancyhead[C]{2022 10th International Conference on Affective Computing and Intelligent Interaction Workshops and Demos (ACIIW)}
\fancyfoot[L]{978-1-6654-5490-2/22/\$31.00 \copyright 2022 IEEE}

\begin{abstract}
Affective robotics research aims to better understand human social and emotional signals to improve human-robot interaction (HRI), and has been widely used during the last decade in multiple application fields. Past works have demonstrated, indeed, the potential of using affective robots (i.e., that can recognize, or interpret, or process, or simulate human affects) for healthcare applications, especially wellbeing.
This paper systematically review the last decade (January 2013 - May 2022) of HRI literature to identify the main features of affective robotics for wellbeing. Specifically, we focused on the types of wellbeing goals affective robots addressed, their platforms, their shapes, their affective capabilities, and their autonomy in the surveyed studies. Based on this analysis, we list a set of recommendations that emerged, and we also present a research agenda to provide future directions to researchers in the field of affective robotics for wellbeing. 
\end{abstract}


\section{Introduction}


The number of people with wellbeing related concerns has increased in the last decade. In addition, the current COVID-19 pandemic has exacerbated this growth leading to societal changes (such as social isolation and work-from-home arrangements) that have severely impacted mental and physical wellbeing. This has resulted in a more urgent need to support people’s wellbeing. 

Affective robotics is a promising venue to support people and help improve their wellbeing. Affective robots can recognize human emotions and show affective
behaviors \cite{rhim2019investigating}, key factors for a successful interaction to promote human wellbeing.
Also, past works have largely used affective robots to improve and maintain both mental (e.g., to aid the evaluation of children's wellbeing related concerns \cite{nida2022evaluation}, cognitive therapy for people with dementia \cite{tapus2009use}), and physical human wellbeing (e.g., to promote exercise activities for the elderly \cite{fasola2013socially}).

However, making an affective robot that is able to recognize, interpret, process and simulate human affect is still an open challenge, because of the several technical challenges (e.g., adaptation to human behavior, personalization of the interaction), and the social and ethical implications of developing and deploying such robots. 

This paper aims at investigating the current state of the art of affective robots for wellbeing.
Specifically, our main research questions is: \textit{"How have affective robots been used to promote people's wellbeing, and to what extent are their affective capabilities suitable to promote wellbeing?"}. We focused on the following sub-questions:
\begin{itemize}
    \item RQ1. What are the wellbeing goals of affective robots?
    \item RQ2. What are the affective robots's platforms (e.g., Nao, Pepper) that have been used for wellbeing?
    \item RQ3. What are the shapes (humanoid, non-humanoid, animal-like) of the affective robots for wellbeing?
    \item RQ4. What are the affective capabilities (e.g., emotion recognition) that the affective robots for wellbeing are endowed with?
    \item RQ5. What are the levels of autonomy (non-autonomous, semi-autonomous,  autonomous) of the affective robots for wellbeing?
\end{itemize}

To answer our research questions, we run a scoping literature review following the PRISMA schema \cite{moher2009preferred} to avoid any bias in the identification, screening, eligibility or inclusion phases. We reviewed the last decade (from January 2013 to May 2022) of HRI literature to provide a detailed picture of affective robotics for wellbeing field.

This paper contributes the following:
\begin{enumerate}
    \item we provide the community with a scoping review of the last decade of HRI works in the affective robotics for wellbeing field;
    \item from the data synthesized, we formulate a list of recommendations for future research in affective robotics;
    \item we identify a research agenda for the affective robotics research field to promote wellbeing.
\end{enumerate}

The rest of the paper is structured as follows.
First, Section \ref{sec:def} defines affective robotics. Next, Section \ref{sec:method}, \ref{sec:results}, and \ref{sec:discussion} are dedicated to the scoping review; describing the methodologies, presenting the main findings, and discussing those findings respectively. We then present our limitations and future works in Section  \ref{sec:lim}  and conclude the paper in Section \ref{sec:concl}.

\section{Definition of Affective Robotics And Challenges}
\label{sec:def}

With the term affective robotics, previous work has referred to the use of affective computing in human-robot interaction. In fact, authors in \cite{rhim2019investigating} defined affective robots as "robots that can recognize human emotions and show affective behaviors". Also, in \cite{churamani2020continual}, they claimed that affective robotics focus on "understanding
human socio-emotional signals to enhance HRI".
Paiva et al. \cite{paiva201421} defined the affective loop of emotional robots as composed of emotion adaptation, emotion expression, and emotion synthesis. However, we acknowledge that having a robot that is able to adapt, express, and synthetise emotions is still difficult, due to several open challenges.
First, affective robotics research has to understand the fundamental mechanisms of human behaviour in real-life circumstances - including nonverbal behavioural cues - and model these  for human-inspired behaviours in robots. 
Second, robots are expected to dynamically adapt to human behaviour, meeting the needs of each individual and personalising their behavior accordingly. 
Third, although affective robots take advantage of the advances in affective computing, the generalisation of those results into real-world context is not straightforward because of the controlled settings usually adopted for creating datasets that inform those affective models. All those challenges make it even more difficult to design affective and intelligent social robots that can support people in promoting their wellbeing. 

Alongside these technical challenges, researchers must also consider the social and ethical implications of developing and deploying such robots, at both the individual and societal levels. 

In this survey, we refer to the term affective robots as \textit{robots that can recognize, interpret, process, or simulate human affect}.

\section{Method}
\label{sec:method}

\begin{figure}[htbp]
\centering
\includegraphics[width=\columnwidth]{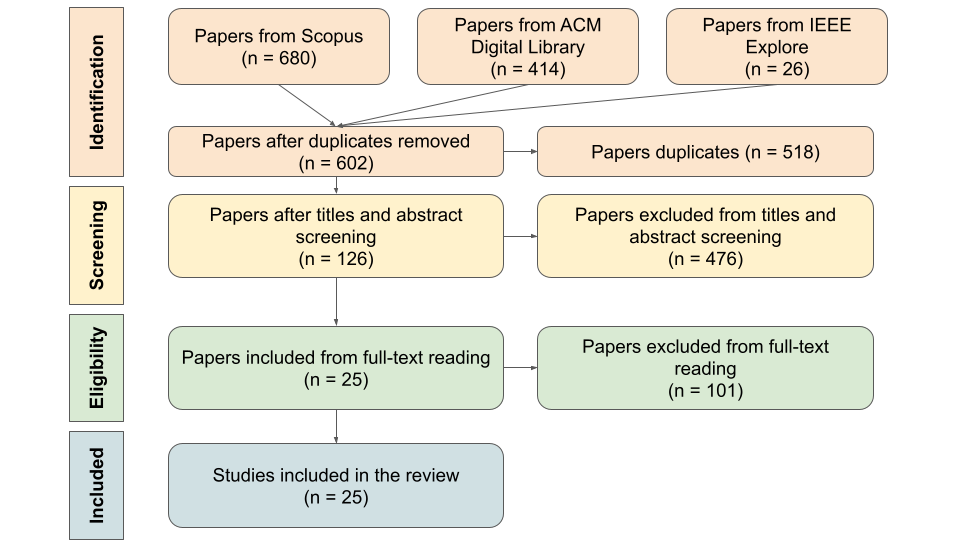}
\caption{PRISMA schema for this survey.}
\label{fig:prisma}
\end{figure}

This survey aims to understand how affective robots have been used to promote people's wellbeing. To address this research question, we followed the PRISMA schema \cite{moher2009preferred} to identify, screen, select, and include the surveyed papers. The steps of this process are shown in Fig.~\ref{fig:prisma}.

\subsection{Search Query}

We identified the surveyed papers searching in the ACM Digital Library, IEEE Explore, and Scopus databases. 
To identify the search query, we exploited the SPIDER \cite{cooke2012beyond} framework. We used similar queries for different databases (i.e., they differ only for the database's search requirements). For the sake of clarity, we present an example of the Scopus search query as follows:

\begin{verbatim}
TITLE-ABS-KEY ( ( "affective robotic*"  OR  
"social robot*"  OR  "emotional robot*" OR 
"socially assistive robot*" )  AND  
( "wellbeing"  OR  "well-being"  OR 
"mental health"  OR  "health" ) )  
AND  PUBYEAR  >  2012
\end{verbatim}

We collected the papers to review by searching via queries in the databases selected, then we removed the duplicates and stored all the resulted references into a CSV file.

\subsection{Eligibility Criteria}
\label{sec:elcrit}

We followed the guidelines of \cite{kitchenham2004procedures} for selecting papers in engineering to identify the set of inclusion and exclusion criteria for this survey.

We included the papers that:
\begin{itemize}
    \item address wellbeing or health (both physical - defined as "mantaining healthy quality of life", such as eating well, exercising, getting enough sleep, staying hydrated - and mental);
    \item employ a physical robot with affective or emotional capabilities (i.e., affective robot);
    \item model or analyse affective capabilities of a robot;
    \item have their title, abstract, and keywords containing at least one keyword describing such technology and one keyword from the Search Query Keywords.
\end{itemize}

We excluded the papers that:
\begin{itemize}
    \item were published before January 2013 and after the day of actual running of the research, i.e., May 16, 2022;
    \item are not in English;
    \item employ a virtual robot (e.g., in VR, or on mobile-based applications);
    \item do not involve human-robot interaction in any form (e.g., running study, analysis data of HRI);
    \item are not in peer-reviewed journals and conference proceedings;
    \item are surveying another topic or theoretical papers;
    \item are inaccessible to the authors;
    \item do not report the details necessary to evaluate their eligibility.
\end{itemize}

\subsection{Selection Process}
We first screened the papers collected at a high level, and then we selected the ones to include in this survey after in-depth analysis.
One reviewer screened the titles and abstracts based on the eligibility criteria listed in Section \ref{sec:elcrit}. The full text of the remaining papers was analyzed and assessed by two reviewers.

\subsection{Data Extraction and Analysis}

\begin{table}[htbp]
\caption{Correspondence between the research questions and variables, types of variable for the extracted data. }
\begin{center}
\begin{tabular}{|c|c|c|}
\hline
\textbf{Research} & \textbf{Variables}                                       & \textbf{Type} \\
\textbf{ Questions}&& \\
\hline
RQ1 & Robot's Goal (e.g., assess/train mental                              & Qualitative  \\
& wellbeing)    & \\
\hline
RQ2                            &               Robotic Platform (e.g., NAO, Pepper) & Categorical \\ 
\hline
RQ3  &   Robot's Shape (e.g., humanoid)                                & Categorical   \\ 

\hline
RQ4   & Robot's Affective Capabilities & Qualitative    \\ 
\hline
RQ5     & Robot's Autonomy (e.g., non- or semi-                                   & Categorical   \\
&or fully- autonomous ) & \\
\hline
\end{tabular}
\label{tab:variable}
\end{center}
\end{table}

To extract data from the surveyed papers, we assigned a variable to each of the five research questions. The types of the variables were either categorical (e.g., robot's autonomy) or qualitative (e.g., robot's goal). Tab. \ref{tab:variable} collects the variables assigned to each research question.
For the categorical data, we defined a priori the classes based on previous literature, e.g., for the robot's autonomy variable we identified three classes - non-autonomous, semi-autonomous, and fully-autonomous - based on \cite{beer2014toward}. 
While for the qualitative data we exploited pattern-based method to extract main themes of the surveyed papers. 

\section{Results}
\label{sec:results}

Following the PRISMA schema, 25 papers were included in this review. The following sections collect the data synthesized and the corresponding research questions addressed. Tab. \ref{tab:results} collects the survey results.


\begin{table*}[]
\label{tab:results}
\caption{Results from the 25 surveyed papers collected by research questions. PW: physical wellbeing, MW: mental wellbeing}
\begin{tabular}{cccccc}

\hline
\textbf{Ref} & \textbf{RQ1}    & \textbf{RQ2} & \textbf{RQ3} & \textbf{RQ4} & \textbf{RQ5} \\
\hline
\cite{erel2022enhancing}           & Emotional Support (MW)                 & Hoffman robot                     & Non-humanoid                     & Movement                        & Autonomous                          \\
\cite{tulsulkar2021can}           & Cognitive Stimulation (MW)         & Nadine                            & Humanoid                         & Emotion Recognition             & Autonomous                          \\
\cite{akiyoshi2021robot}          & Emotional Support (MW)                 & Sato                              & Humanoid                         & Semantic Understanding          & Autonomous                          \\
\cite{taylor2021exploring}         & Cognitive Stimulation (MW)           & Stevie                            & Humanoid                         & Facial Expressions              & Semi-autonomous                     \\
\cite{bodala2021teleoperated}           & Mindfulness (MW)                        & Pepper                            & Humanoid                         & Facial Expressions and Movement & Non-autonomous                      \\
\cite{bjorling2020exploring}           & Self-disclosure (MW)                    & Darwin mini                       & Humanoid                         & Semantic Understanding          & Autonomous                          \\
\cite{avioz2021robotic}        & Physical Stimulation (PW)               & Nao and Poppy                     & Humanoid                         & Facial Expressions and Movement & Semi-autonomous                     \\
\cite{do2020clinical}          & Assessment Via  & 3D printed robot                  & Humanoid                         & Emotion Recognition             & Autonomous                          \\
          & Clinical Interviews (MW) &               &                       &            &                           \\
\cite{spekman2021physical}          & Emotional Support (MW)                 & Nao                               & Humanoid                         & Facial Expressions and Movement & Non-autonomous                      \\
\cite{zhang2020socially}          & Physical Stimulation (PW)              & Cozmo                             & Humanoid                         & Facial Expressions and Movement & Autonomous                          \\
\cite{jeong2020robotic}          & Emotional Support (MW)                 & Jibo                              & Humanoid                         & Movement                        & Autonomous                          \\
\cite{rossi2020emotional}          & Emotional Support (MW)                 & Nao                               & Humanoid                         & Facial Expressions and Movement & Semi-autonomous                     \\
\cite{bjorling2020exploring}          & Emotional Support (MW)                 & Emarv4 and Blossom                & Non-humanoid and Animal-like     & Facial Expressions and Movement & Non-autonomous                      \\
\cite{miyake2020towards}          & Fall Detection (PW)                     & Side-bot                          & Humanoid                         & Facial Expressions              & Autonomous                          \\
\cite{deublein2020expressive}          & Physical Stimulation (PW)             & Reeti                             & Non-humanoid                     & Facial Expressions and Movement & Non-autonomous                      \\
\cite{triantafyllidis2019social}          & Food Promotion (PW)                    & Cozmo                             & Non-humanoid                     & Facial Expressions and Movement & Semi-autonomous                     \\
\cite{shao2019you}          & Physical Stimulation (PW)              & Pepper                            & Humanoid                         & Emotion Recognition             & Autonomous                          \\
\cite{rhim2019investigating}          & Emotional Support (MW)                 & Pepper                            & Humanoid                         & Facial Expressions and Movement & Semi-autonomous                     \\
\cite{feng2019livenature}          & Emotional Support (MW)                 & IRS                               & Animal-like                      & Facial Expressions and Movement & Semi-autonomous                     \\
\cite{block2019softness}         & Emotional Support (MW)                 & PR2                               & Not-humanoid                     & Movement                        & Semi-autonomous                     \\
\cite{ritschel2018drink}          & Food Promotion  (PW)                   & Reeti                             & Non-humanoid                     & Facial Expressions              & Autonomous                          \\
\cite{spekman2018perceptions}          & Emotional Support (MW)                 & Alice                             & Humanoid                         & Facial Expressions and Movement & Semi-autonomous                     \\
\cite{jeong2018huggable}          & Emotional Support (MW)                & Huggable                          & Animal-like                      & Facial Expressions and Movement & Non-autonomous                      \\
\cite{johnson2016exploring}          & Entertainment (MW)                    & Nao                               & Humanoid                         & Facial Expressions and Movement & Semi-autonomous                     \\
\cite{magyar2015socially}          & Physical Stimulation (PW)              & Nao                               & Humanoid                         & Facial Expressions and Movement & Non-autonomous         \\
\hline

\end{tabular}
\end{table*}

\subsection{Affective Robot's Goal (RQ1)}

\begin{figure}[htbp]
\centering
\includegraphics[width=\columnwidth]{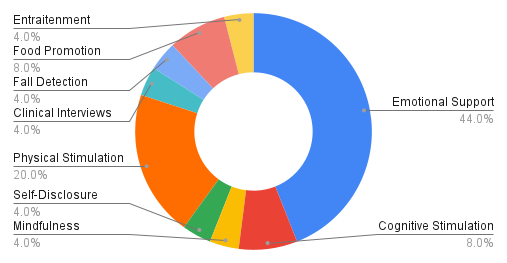}
\caption{Affective robot's goal (RQ1) in the surveyed papers among January 2013 to May 2022}
\label{fig:rq1}
\end{figure}
Overall, 17 studies used affective robots to promote mental wellbeing, in 8 studies the robot was used for physical wellbeing. Fig. \ref{fig:rq1} details the robot's goal, and specifically, we identified 9 different goals for affective robots that promote wellbeing among the surveyed papers. 
Eight studies focused on physical wellbeing. Five of the studies (20\%) focused on physical stimulation to promote physical wellbeing. Two studies (8\%) explored the use of affective robots to promote healthy food (or drink) for physical wellbeing. One study (4\%) used the affective robot to detect fall. Then, the remaining 17 studies focused on promoting mental wellbeing. 11 studies (44\%) aimed to provide an emotional support - including reducing stress or anxiety - to promote mental wellbeing of the participants. 2 (8\%) investigated affective robots to facilitate cognitive stimulation promoting mental wellbeing. 1 of the studies (4\%) aimed at promoting self-disclosure by using an affective robot. One study (4\%) adopted affective robots to assess participants via clinical interviews. One study (4\%) used the robots to entertain participants. Finally, one study (4\%) exploited the affective robot to provide a mindfulness session. 

To sum up, to date the HRI community focused more on the application of affective robotics to promote mental wellbeing. Some of them exploited affective robots to facilitate physical wellbeing as well. 

\subsection{Affective Robot's Platform (RQ2)}

\begin{figure}[htbp]
\centering
\includegraphics[width=\columnwidth]{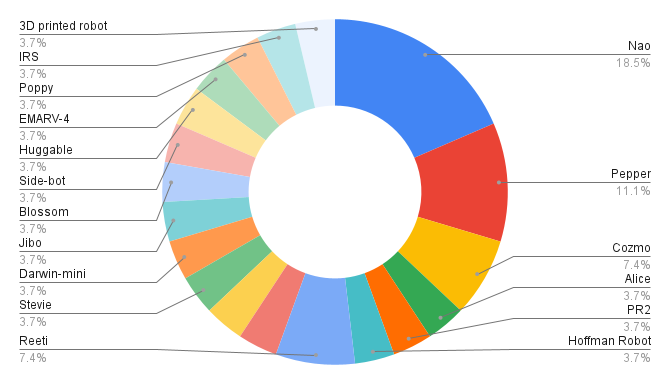}
\caption{Affective robot's platform (RQ2) in the surveyed papers among January 2013 to May 2022}
\label{fig:rq2}
\end{figure}

From the surveyed studies, we identified 17 different robotic platform, as depicted in Fig. \ref{fig:rq2}. Specifically, 5 of the studies (19.5\%) used a Nao robot, 3 studies (11\%) Pepper, 2 studies (7.5\%) Cozmo, 2 studies (7.5\%) Reeti, one study (3.7\%) Alice, one study (3.7\%) PR2,  one study (3.7\%) a non-humanoid robot by Hoffman \cite{hoffman2015design},  one study (3.7\%) Stevie,  one study (3.7\%) Darwin-mini,  one study (3.7\%) Blossom,  one study (3.7\%) Jibo,  one study (3.7\%) Side-bot,  one study (3.7\%) Emarv-4,  one study (3.7\%) IRS,  one study (3.7\%) Huggable, and  one study (3.7\%) a 3D printed social robot.

Those results show that the HRI community is gradually exploring different robotics platforms that are now more available in the market. Still, the robotics platforms from SoftBank (Nao and Pepper) are the most commonly used within the HRI community. 

\subsection{Affective Robot's Shape (RQ3)}
\begin{figure}[htbp]
\centering
\includegraphics[width=0.83\columnwidth]{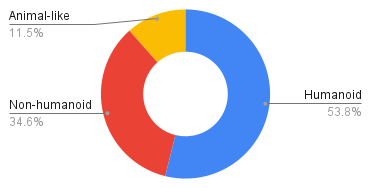}
\caption{Affective robot's shape (RQ3) in the surveyed papers among January 2013 to May 2022}
\label{fig:rq3}
\end{figure}

To cluster the affective robot's shape of the surveyed papers, we used the definition by \cite{fong2003survey} and \cite{shibata2004overview}, who defined the robot agent shape as bio-inspired (e.g., animal-like, humanoid), and non bio-inspired (artificial, i.e., showing artificial characteristics, and functional, i.e., performing specific functional tasks).

Among the survey papers, 16 out of 26 (65\%) used a bio-inspired robot (note that one study \cite{bjorling2020exploring} adopted two robotics platforms).
Particularly, 14 studies (54\%) adopted humanoid affective robots, while 3 studies (11\%) employed animal-like affective robots.
The remaining studies (9 out of 26, 35\%) adopted affective robots with non-humanoid shape. Fig. \ref{fig:rq3} depicts the results of the affective robot's shape for the surveyed papers.

In summary, those findings show that HRI researchers opted for affective robots with bio-inspired shape (specifically humanoids), just a few of them investigated the non bio-inspired shapes.

\subsection{Affective Capabilities (RQ4)}

\begin{figure}[htbp]
\centering
\includegraphics[width=0.7\columnwidth]{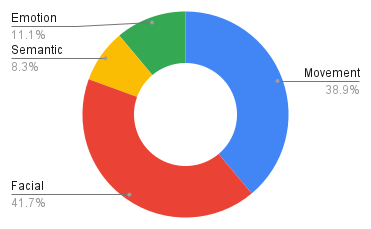}
\caption{Affective robot's capabilities (RQ4) in the surveyed papers among January 2013 to May 2022}
\label{fig:rq4}
\end{figure}

We clustered the affective capabilities of the robots included in the surveyed papers into four main classes (see Fig. \ref{fig:rq4}): emotion recognition, semantic emotional understanding, facial expressions, and emotional movements.

11 out of the surveyed studies endowed the robot with both facial expression and emotional movement capabilities. Particularly, 14 studies (39\%) equipped the robot with the capability of expressing emotional movements, while 15 out of 25 (42\%) endowed the robots with the capability of expressing emotion through facial expressions. 4 studies (11\%) adopted robots that were able to automatically recognize emotions in participants. Finally 3 studies (8\%) used robots that were able to semantically understand the participants' emotions.

To sum up, our findings show that the HRI community has mainly focused on endowing their robots for wellbeing with the capability of expressing emotions (via facial expressions or movements). More recently, the HRI researchers started to equip their robots with automatic user affect detection capabilities to promote wellbeing.

\subsection{Affective Robot's Autonomy (RQ5)}
\begin{figure}[htbp]
\centering
\includegraphics[width=0.9\columnwidth]{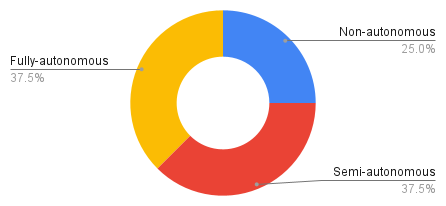}
\caption{Affective robot's autonomy (RQ5) in the surveyed papers among January 2013 to May 2022}
\label{fig:rq5}
\end{figure}
To identify the level of autonomy of the robots, we followed the definition provided by \cite{beer2014toward}, in case the authors did not provide any detailed specifications.

Among the surveyed studies, 6 of them (25\%) adopted a non-autonomous affective robots where the researcher acted as "wizard" in the interaction (aka, Wizard-of-Oz method). 9 of the studies (37.5\%) exploited a semi-autonomous affective robot. For example, in \cite{taylor2021exploring}, the robot could automatically execute the task interaction, but the research controlled its navigation in the care center. The other 9 studies (37.5\%) adopted a fully autonomous affective robot (see Fig. \ref{fig:rq5}). 

Our findings showed that the HRI community is moving towards more autonomous robots, despite some researchers preferring to exploit tele-operated robots (i.e., non-autonomous) to better control the design variables in their studies.

\section{Discussion}
\label{sec:discussion}
The next sections will discuss the results gathered in this survey. Specifically, we extrapolated a set of recommendations that we list in Section \ref{sec:rec} and a research agenda in Section \ref{sec:ra}

\subsection{Recommendations}
\label{sec:rec}
We list a set of observations and subsequent recommendations as follows.

First (from RQ1), we found that the surveyed papers exploited affective robots to promote either mental (e.g., \cite{akiyoshi2021robot, devillers2021human, tulsulkar2021can}) and physical wellbeing (e.g., \cite{avioz2021robotic, zhang2020socially}). Specifically, we observed that affective robots that addressed mental wellbeing adopted both expressive behaviors (e.g., facial expression, emotional movements) and emotion recognition capabilities (e.g., semantic understanding), while the affective robots for physical wellbeing mostly were endowed with expressive capabilities only. Also, our results showed that the authors who focused on affective robots for physical wellbeing opted mostly for human-like shape. The humanoid shape is extremetely important when delivering physical exercise, so that human participants can replicate the robot movements, as in \cite{avioz2021robotic}. On the other hand, affective robots for mental wellbeing used both humanoid and non-humanoid robots to deliver tasks that can aid the emotional support \cite{bjorling2020exploring} or the cognitive stimulation \cite{taylor2021exploring}. 
Finally, most of the papers on physical wellbeing chose an automated or semi-automated affective robot, while the one that focused on mental wellbeing exploited both autonomous and non-autonomous robots with Wizard-of-Oz approach

Second (from RQ2 and RQ3), we found that most of the surveyed papers adopted a bio-inspired robot form, especially humanoid (e.g., \cite{rhim2019investigating, shao2019you}). Past works demonstrated that the human-like appearance of the robot influences the expectations of the users. For example, participants attributed human-like behaviors to humanoid robots, because they associated their shape with their functionalities \cite{axelsson2021participatory}. On the other hand, the remaining papers in the review adopted a non-humanoid shape. 
In fact, a previous work \cite{axelsson2021participatory} showed that participants who were asked to rank different robotic platforms (e.g., Jibo, Pepper, Miro) provided contradictory responses. Some of the participants preferred more humanoid robot shape, while other chose an abstract shape as the most suitable to be a robotic coach delivering wellbeing interventions.

Then (from RQ4), our results showed that the affective capabilities of the robots focused more on generating expressive behaviors (e.g., \cite{deublein2020expressive, feng2019livenature}). Just few of them (e.g., \cite{rossi2020emotional}) displayed capabilities of affect detection. In the context of promoting wellbeing, having a robot that is endowed with the capabilities of both affect synthesis and affect detection is a key factor for a successful interaction. In fact, in \cite{axelsson2021participatory}, the authors reported that participants, in a participatory design study, remarked the need to endow the wellbeing robot coach with both the capabilities of emotion/empathy generation  and emotion detection to be able to adapt to users.

Finally (from RQ5), we found that most of the surveyed papers adopted autonomous or semi-autonomous affective robots, that is likely because of the advance in the fields of natural language processing \cite{chowdhary2020natural}, computer vision \cite{khan2021machine}, and speech recognition \cite{nassif2019speech} in the last decade. Despite the increasing interest in autonomous robots, many researchers exploited non-autonomous robots that are directly controlled or tele-operated by a human operator. This method allows researchers to overcome technical issues, that can compromise the interaction and the user's perception, and unpredictable events, typical of user studies. \\

As a result of these observations we provide the following four recommendations:

\begin{itemize}
    \item \textbf{Recommendation 1 -} We recommend to use autonomous humanoid robots to deliver exercises that promote physical wellbeing to enable users to imitate the robot movement, and endow the robots with detection capabilities to check the user's affective state during the activity. To promote mental wellbeing, we suggest that the researchers design the affective robot as autonomous and endowed with both affect expression and  detection capabilities. Both humanoid and non-humanoid shape can be exploited depending on the task or exercise that the robot has to deliver.
    \item \textbf{Recommendation 2 -} We recommend to choose the shape of the affective robot to employ in the study according to its functionalities. For example, in physical rehabilitation, it could be more useful to use a humanoid robot that can display movement to perform physical exercises, while for reducing loneliness in elderly, an animal-like shape could be more appropriate to resemble the function of a pet-companion. Both humanoid and non-humanoid robots seem to be appropriate for delivering mental wellbeing exercises.
    \item \textbf{Recommendation 3 -} We then recommend to endow the affective robots with both capabilities of generating expressions and detecting the affective state of their user to be able to deliver wellbeing exercises in more effective and adaptive ways.
    \item \textbf{Recommendation 4 -} We recommend to lean toward the autonomous affective robots to advance further the field of robotics and provide evidence of the efficacy of real robots in the real world.
\end{itemize}

\subsection{Research Agenda}
\label{sec:ra}
Future research should focus on the design features of affective robots to promote wellbeing. Specifically, much work needs to be done to investigate which robot form is more appropriate for which specific task (e.g., physical exercise, cognitive stimulation) to guide researchers in the design choice of the robotic platform. For instance,  \cite{haring2018ffab} demonstrated that humans have a form function attribution bias which affects their perception of the robots. People take a cognitive shortcut to attribute the functionality of the robot using the visual information.  Within the HRI literature, many efforts have been made to better understand how the robot's form - in terms of size, gender, and appearance - affects the user's perception of the robot \cite{li2015robot, haring2018ffab, schaefer2012classification}. However, future research should specifically focus on better understanding how form impacts user perception of the robots utilised for wellbeing related applications, and how robot form can make a difference in the efficacy of the delivered intervention. 

Then, we believe that future research should focus on empirical studies that adopt affective robots endowed with the full spectrum of capabilities, i.e., to recognize, understand, and generate expressions, to advance the affective robotics field, providing also evidence on the current technology readiness level for the real world.  Deploying social robots in real human-robot interaction settings is still an open challenge \cite{park2020towards} mainly due to the need for real-time processing capabilities and the lack of computational power of the robotic platforms available in the market. 
The lack of cross-fertilization between affective computing and social robotics fields also contributes to this problem \cite{celiktutan2018computational}. Future efforts should focus on how to overcome those technological limitations, for example, using cloud computing, or using external/enviromental sensors, as suggested in \cite{celiktutan2018computational}. 

Finally, affective robotics should lean towards creating and/or using autonomous robots. To this end, we acknowledge that future research should address some of the technological limitations related to robotic deployment \cite{devillers2021human}. In parallel to the advances in computing power, the field of affective computing has seen a rapid progress, however, there is still a lack of real-time studies with robots endowed with, for example, affect recognition capability that can adapt and personalize to each user during the interaction \cite{churamani2020continual}. A survey on 10 years of HRI studies \cite{riek2012wizard} showed that the most common types of Wizard control employed were natural language processing and non-verbal behaviors, including affective capabilities of the robot. Also, the Wizard-of-Oz technique has been widely used in the HRI community \cite{ belpaeme2020advice}, because deploying autonomous robots is an open problem. In fact, the main challenges are:
\begin{enumerate}
    \item programming autonomous social behavior for a robot is very difficult and time consuming;
    \item researchers usually program human-robot interactions as a one off experience, for a limited scope and very short interaction durations (usually no longer than 20 minutes);
    \item the off-the-shelf robotic platforms usually fail in meeting the user expectations in terms of robot capabilities
\end{enumerate}

To overcome those limitations, the next generation of HRI works should focus on how to make robots more autonomous using data-driven approaches to fully understand the dynamics of human and autonomous robot interactions.

\section{Limitations and Future Work}
\label{sec:lim}
One of the main limitations of this survey is that the screening of the papers have been conducted by a single reviewer. This could have introduced bias into the paper inclusion. We plan to involve at least one additional researcher to provide a more solid method to include the  papers to review.
Another issue is that the set of research questions explored are limited. There are several aspects that we have not covered in this paper that are known to impact HRI, such as context \cite{wang2019investigating}. 
With a deeper analysis, we can identify missing points relevant for the HRI community that can further inform our research agenda. The recommendations of this paper were also mentioned previously in different articles in the literature but we haven't reported all of them. Moreover, many of these recommendations are not based on the findings of the review, but we grounded them on literature. Finally, in the recommendations we haven't mentioned the difficulties involved in developing the affective robots that could have impacted the works of the reviewed papers.
In our future work, we will extend this survey with a broader set of research questions (e.g., study design, participant population etc.) addressing the above-mentioned limitations.

\section{Conclusions}
\label{sec:concl}

This paper reviewed the last decade (2013-2022) of HRI literature on affective robotics for wellbeing utilising the PRISMA schema.
We aimed to understand how affective robots have been used in previous studies to promote human wellbeing and to what extent their affective capabilities were useful.
Our findings showed that the HRI community: i) focused mostly on affective robots to promote mental wellbeing, ii) explored different robotic platforms, iii) opted for human-like affective robots, iv) endowed  robots mostly with the capability of generating expressive behaviors, and v) adopted mostly autonomous and semi-autonomous robots.  
The results from this review enabled us to list a set of recommendation guidelines and a research agenda for future research in the affective robotics field.

\section*{Ethical Impact Statement}
We acknowledge that our paper did not survey the ethical implications of using affective robots. However, this is out of the scope of this paper. In our future work, we will address also ethical concerns in the field of affective robotics that aims to promote wellbeing.

\section*{Acknowledgment}
 This work is supported by the EPSRC project ARoEQ under grant ref. EP/R030782/1. For the purpose of open access, the authors have applied a Creative Commons Attribution (CC BY) licence to any Author Accepted Manuscript version arising.

\bibliographystyle{IEEEtran}
\bibliography{ACII2022-template}

\end{document}